\RequirePackage{fix-cm}

\documentclass[twocolumn]{svjour3}        
\smartqed  

\usepackage[pdftex]{graphicx}
\usepackage{float}
\usepackage{multirow}
\usepackage[export]{adjustbox}
\usepackage{figsize}
\usepackage{lipsum,mdframed,xcolor}
\usepackage{tikz}
\usepackage{amssymb}
\usepackage{pifont}
\newcommand{\xmark}{\ding{55}}%
\usepackage{subfigure}
\usepackage{algorithmic}
\usepackage[]{algorithm2e}
\usepackage[font=footnotesize]{subfig}
\usepackage{diagbox}
\usepackage[outline]{contour}
\usepackage{stfloats}
\usepackage{authblk}
\usepackage{tabularx}
\usepackage[numbers,sort&compress]{natbib}

\usepackage[hyphens]{url} 
\definecolor{mycolor}{rgb}{0.9, 0.1, 0.1}
\usepackage[british]{babel}
\def\makeheadbox{\relax}

\begin{document}
\sloppy
\title{A limited-size ensemble of 
homogeneous CNN/LSTMs for
 high-performance word classification
}

\author{Mahya Ameryan 
\and
 Lambert Schomaker}

\institute{M. Ameryan 
             \\
              \email{m.ameryan@rug.nl}             \\
           L. Schomaker %
\\
             \email{l.r.b.schomaker@rug.nl}
\\
             Artificial Intelligence and Cognitive Engineering, Faculty of Science and Engineering, University of Groningen, Groningen, The Netherlands
}

\maketitle

\begin{abstract}
\ In recent years, long short-term memory neural networks (LSTMs) have been applied quite successfully to problems in handwritten text recognition. 
However, their strength is more located in handling sequences of variable length than in handling geometric variability of the image patterns. 
Furthermore, the best results for LSTMs are often based on large-scale training of an ensemble of network instances. 
In this paper, an end-to-end convolutional LSTM Neural Network is used to handle both geometric variation and sequence variability.
 We show that high performances can be reached on a common benchmark set by using proper data augmentation for just five such networks using a proper coding scheme and a proper voting scheme.
 The networks have similar architectures (Convolutional Neural Network (CNN): five layers, bidirectional LSTM (BiLSTM): three layers followed by a  connectionist temporal classification (CTC) processing step). 
The approach assumes differently-scaled input images and different feature map sizes.
Two datasets are used for evaluation of the performance of our algorithm: A standard benchmark RIMES dataset (French), and a historical handwritten dataset KdK (Dutch).
 Final performance obtained for the word-recognition test of RIMES was 96.6\%, a clear improvement over  other state-of-the-art approaches. On the KdK dataset, our approach also shows good results. 
The proposed approach is deployed in the Monk search engine for historical-handwriting collections. 

\end{abstract}
\section{Introduction}

Convolutional neural networks (CNNs) \cite{LeCun:1998:CNI} and long short-term memory networks (LSTM) \cite{Hochreiter:1997:LSTM} and its variants \cite{Graves2005bilstm,Graves2009} have recently achieved impressive results \cite{Li2014ConstructingLS,Gideon2019,Doetsch2014ney}. This exceptional performance comes, however, at the cost of having an ensemble of, e.g., 118 recognizers \cite{Stuner2016HandwritingRU}. High cost of training and operation brings to mind the question whether less costly methods can be applied to boost the performance of handwriting recognizers. 
 
\begin{figure}[t!]
\centering
\includegraphics[scale=0.25]{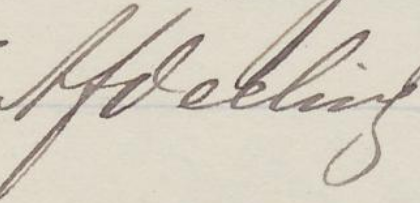}
\caption{ A historical spelling of a word, \textit{Afdeeling}, in the historical KdK dataset. The contemporary spelling of this word would be \textit{Afdeling}. }
\label{fig:afdeeling}
\end{figure}
A possible direction would consist of the use of linguistic statistics \cite{Puigcerver2018}. A recent method for using language information is a dual-state word-beam search \cite{Scheidl2018} for decoding the connectionist temporal classification (CTC \cite{Graves:2006:CTC}) layer of neural networks, which has been shown to be effective \cite{Scheidl2018}. 
 
Although the presence of dictionaries and corpora is beneficial, historical documents present a challenge. For instance, historic spelling of a word differs from the contemporary spelling, there often is an absence of strict orthography, and there may be frequent misspellings \cite{hauser2007unsupervised}. Figure \ref{fig:afdeeling} shows a word image from one of the datasets used in this paper. This historical word has an extra character compared to the current spelling. Moreover, for rare languages, e.g., Aymara \cite{Emlen2017}, the  complete lexicon does not exist yet, and corpora are of very limited size. Handwritten-text recognition (HTR) is exactly required to {\em obtain} such digital linguistic resources for that language.

Another possible direction to improve performance  would concern a heavy optimization of network architecture and training (hyper)parameters. 
The state-of-the-art approaches can be sensitive to the choice of hyper-parameter values. As an example, it is reported that increasing the depth of a neural network that consists of convolutional and LSTM layers, from 8 hidden layers to 10 is advantageous. Further enlarging to 12 hidden layers yielded unsatisfactory results \cite{Voigtlaender2016}. From the perspective of e-Science services for handwriting recognition dealing with hundreds different books, it is not feasible to tailor the recognizer models for each book based on prior knowledge, using human handcrafting of neural networks. Preferably, having an ensemble consisting of a limited number of automatically generated architectures would be practical. 
 
In this paper, we explore the possibilities of exploiting the success of current CNN/LSTM approaches, using several methods at the level of linguistics and labeling systematics, as well as an ensemble method. Ideally, the approach should be robust, require a minimum of human intervention with a limited set of hyper-parameter settings (architectures), and minimum linguistic resources. For evaluation, we use a standard benchmark public dataset, RIMES \cite{grosicki:RIMES}, and a historical handwritten dataset, KdK \cite{Zant2008,VANOOSTEN20141031}. The two dataset differ in time period and language.

An essential consideration is that it should be possible to add our suggested algorithm to the Monk system, \cite{Zant2008, Zant2009,sheng2016,Schomaker2019}. Monk is a live web-based search engine for words and character recognition, retrieval and annotation. It contains diverse digitized historical and contemporary handwritten manuscripts in  many languages: Chinese, Thai, Arabic, Dutch, English, Persian. Also, complicated machine-printed documents  such as German, Fraktur, Egyptian hieroglyphs, and historical language are available in the Monk system.  

The rest of this paper is structured as follows. 
In section \ref{RelatedWorksec}, we briefly survey the related works in terms of 
recent state-of-the-art methods on RIMES, convolutional recurrent neural network, word search in character-hypothesis grids, ensemble systems, and requirements of the proposed method. In section  \ref{Methodsec}, we present our system. The experimental evaluation and discussion are given in sections \ref{resultsec} and \ref{discussionsec}. Finally, conclusions are drawn in section \ref{conclusionsec}.
\section{Related Work}\label{RelatedWorksec}
In this section, we first briefly survey the recent studies that worked on isolated words of the RIMES dataset. Then, a convolutional recurrent neural network is briefly detailed. Afterwards, we survey part of long  history of word search in character-hypothesis grids and linguistic post-processing. As an example of these approaches, we explain a dual-state word-beam search for CTC decoding which is one of principle of our work. Finally, we review researches in ensemble-system approaches.

\subsection{On RIMES}
One of the used datasets in this paper is RIMES \cite{grosicki:RIMES}. In this section, the compared methods are explain briefly. In \cite{Poznanski}, a 12-layer convolutional neural networks (CNN) is used to processes fixed-sized word images and recognize a  Pyramidal Histogram of Character (PHOC) representation \cite{Almazán2014}, using multiple parallel fully connected layers. Afterwards, Canonical Correlation Analysis (CCA)\cite{CCA} is applied as a final stage of the word recognition task, using  a predefined lexicon. 

In \cite{Stuner2016HandwritingRU}, two architectures are used to generate more than a thousand networks to construct an ensemble. Each network is either two-layer BiLSTM or three-layer multidimensional LSTM (MDLSTM) neural networks \cite{Graves2009}. BiLSTMs are fed by HOG \cite{Dalal}, and the input of the MDLSTM is raw image. The \textit{best path} algorithm \cite{Graves2012} is applied for CTC decoding. This approach uses a lexicon verification method. After training 2,100 networks and evaluating on the validation set of RIMES dataset, the lowest performance networks are removed, which results in 118 networks. It is reported that the pruned ensemble of 118 networks has 0.16pp drop in performance compared to the ensemble of 2,100 networks on the RIMES dataset. On another dataset, IAM \cite{IAM}, the size of ensemble is different (n$_{rec}$=1,039). Because of the simplicity of system and high number of recognizers, the complexity is medium to high.
 
In \cite{Menasri2012}, an ensemble uses eight recognizers for handwriting recognition which includes four variants of a MDLSTM, a grapheme based MLP-HMM, and two variants of a context-dependent sliding window GMM-HMM. The ensemble system is a simple sum rule.

In \cite{Sueiras2018}, a framework consisting of a deep CNN, LSTM layers as encoder/decoder, and a
attention mechanism for isolated handwritten-word recognition is given. The result is reported with/without dictionary. For pre-processing, methods for baseline correction, normalization, and deslanting are applied. After pre-processing, an input image is converted to a sequence of image patches by using a horizontal sliding window, . Then, a deep CNN  is used for feature extraction. Afterwards, a LSTM is applied to extract the horizontal relationships existing among a sequence of overlapped horizontal patches of input images. Then, a decoder component is used, a combination of an LSTM and an attention mechanism. To find the best performance, experiments are done to determine the optimal LSTM cell size and patch size. This method does not have very high performance.
 
In \cite{Ptucha2019}, a whole-word CNN can be apply to recognize known words, defined as the 500 most frequent words in the training set of the RIMES dataset, which have a minimum confidence level of 70\%. Otherwise, a Block Length CNN predicts the number of symbols in the given image block. Then, a fully convolutional neural network predicts the characters. Finally, the result is enhanced by a vocabulary-matching method. This varied-CNN method has a problem with separating common and non-common words. The separation of lexicon into a set of common and a set of uncommon words may be artificial, in view the usual continuously decaying Zipf distribution \cite{Zipf}. In \cite{Dutta2018}, deslanting and slope normalization is performed on images, using the approach presented in \cite{Vinciarelli2001}. A pre-trained CNN-RNN is used. During training and testing on benchmark datasets, three types of augmentations are used: affine transformation; elastic distortion; multi-scale transformation. Then, the best result on one of their seven approach is reported. Before that, image augmentation during training and testing is used in \cite{Okafor2017,Okafor2018} for animal recognition. 

The successful methods applied to the RIMES dataset are unfortunately quite complicate. Most of them use a combination of CNNs and LSTMs. Therefore, we treat convolutional neural networks in the next section.
\subsection{Convolutional Recurrent Neural Network}

The convolutional recurrent neural network is an end-to-end trainable system presented in \cite{Shi2017AnET}.
It outperforms the plain CNN in three aspects: 1) It does not need precise annotation for each character and it can handle a string of characters for the word image; 2) it works without a strict preprocessing phase, hand-crafted features or component localization/segmentation; 3) It benefits from the state preservation capability of a recurrent neural network (RNN) in order to deal with character sequence; 4) It does not dependent on the width of word image. Only, height normalization is needed.  

The model is composed of seven layers of  convolutional layers followed by two layers of BiLSTM units containing 256 hidden cells and a  transcription layer. Although, a the model is made up of two distinct neural network varieties, it can be trained integrally using one loss function.

\begin{figure}[!b]
\centering
\includegraphics[width=2.5in]{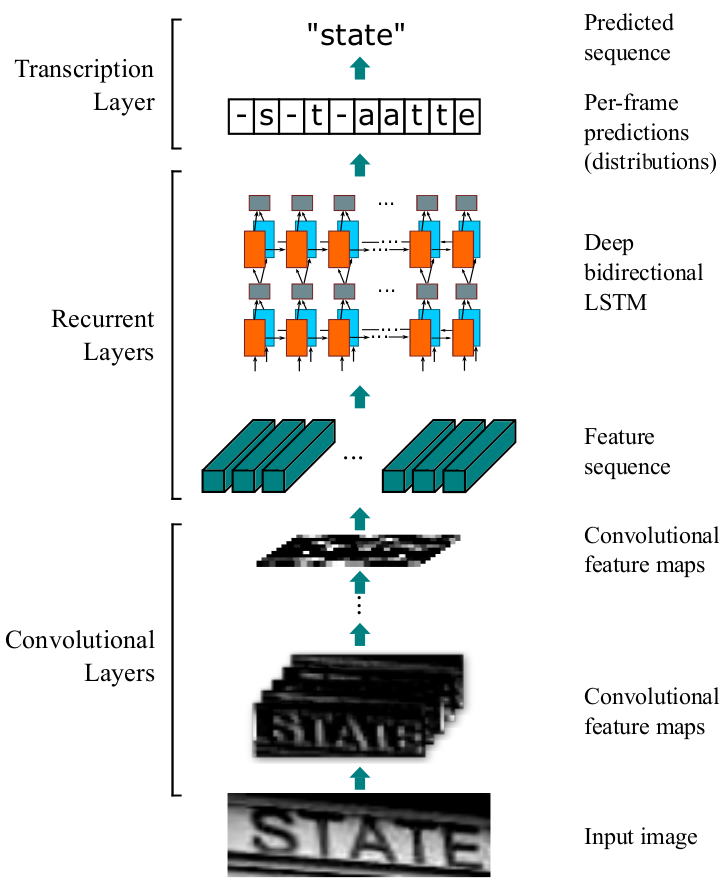}
\label{CRNN-shi2016}
\caption{The architecture of a convolutional recurrent neural network is composed of three components: convolutional, recurrent layers and transcription layer. The phases are as follows: First, feature extraction is carried out by convolutional  layers directly from a height-normalized and gray-scale word image. Secondly, for each frame, prediction of label distribution is performed by RNN layers. Thirdly, the transcription layer transcribes the regarding prediction into a label sequence \cite{Shi2017AnET}.}
\label{fig_sim}
\end{figure}

Figure \ref{fig_sim} shows the pipeline of the convolutional recurrent neural network \cite{Shi2017AnET}.
The input of the model is a height-normalized and gray-scale word image. The the feature extraction is performed by convolutional layers directly from the input image. The output of CNN is a frame of features sequence, and acts the input of the recurrent neural network, which provides raw character hypotheses. Finally, the transcription layer translates the resulting prediction into a label sequence. 

\subsection{Word search and linguistic post-processing}

Character-oriented approaches create a data structure representing the character hypotheses, their position in the text and the confidence value. For example, a LSTM produces a final map with character hypothesis activations, ordered from left-to-right or right-to-left with some stride (step size). Other approaches generate a grid or graph of character hypotheses. The final processing step involves finding the most likely character path, given a dictionary and potential other linguistic resources (statistics). For the LSTM, a well-known first step toward this is connectionist temporal classification (CTC) \cite{Graves:2006:CTC}.

Given a dictionary containing possible input words, an easy method can be used for error detection and correction of a word recognizer. In the case of existence of the word hypothesis in the dictionary, the result is accepted as the label of the input image. Otherwise, if a similar word exists in the dictionary, it can be accepted as final label candidate by using the Levenshtein distance and its variants \cite{Levenshtein1966,Wagner1974,Seni1996,Oommen1997}, or  n-gram distances \cite{Angell1983}, as common measures for comparing $($dis$)$similarity. If required, it is possible to use suitable linguistic statistics to further refine the ranking \cite{Basil2012,Asonov2010,Amrhein2018}.

A data structure for contextual word recognition is presented in \cite{WELLS1990} for quick dictionary look-up using limited memory.

An approach of providing contextual information by giving a dictionary to predict the most probable label in a graph search is presented in \cite{Favata2001}, which is robust to dictionary errors. In this approach, for every lexical word, the most probable path and related confidence is calculated to predict dictionary ranking.

Shannon \cite{Shannon1951} \cite{Shannon1948} was one of the first researchers working on the letter prediction task. Based on this idea, using a trainable variable memory length Markov model (VLMM), a linguistic post-processing model for character recognizers is introduced in \cite{Guyon1996}. The next character is predicted by a variable length window of previous characters.

In \cite{Swaileh2018}, on
the linguistic corpora, a statistical n-gram language model of syllables is trained. In \cite{Liu2002}, for Japanese mail address, a character recognition method uses a dictionary in a trie tree. The dictionary matching is controlled by a beam search approach. The dictionary includes all the address names and principal postal offices in Japan. After pre-processing and segmentation character hypotheses are produced by combination of successive segments. Then, a version  of a nearest-neighbor classifier that exploits the trie structure is made for a fast predict in of the final label. 
In \cite{Seni1996-2}, an on-line handwritten recognition system for cursive words uses simple character features to reduce a given large dictionary. The outputs of  a Time-Delay Neural Network (TDNN) are converted into a character sequence. The result of the system is a matched word in the reduced dictionary using  a variant of Damerau-Levenshtein distance.
For on-line handwriting recognition a search technique is proposed in \cite{seni2000}, which is a post-processing phase of a recognition system that calculates posterior probabilities of characters based on Viterbi decoding. 

In \cite{seni1999} a version of beam and Viterbi search-recognizer is presented. This search method provides the use of discrete probabilities generated by many character recognition systems based on stroke. \cite{Powalka1993} introduces a technique combining word segmentation and character recognition with lexical search to deal with segmentation ambiguities. A depth first trace of dictionary tree for text recognition using recursive procedure presented in \cite{Ford1990}. For online handwriting recognition, in \cite{Bramall1995}, by applying simple feature extraction a given dictionary is reduced. Afterwards, the reduced dictionary is refined by AI techniques. In \cite{Côté1998}, for isolated  cursive handwritten-word recognition, contextual knowledge is used. A dictionary tree representation with efficient pruning method, as a fast search method for large dictionary for on-line handwriting recognition system is proposed in \cite{Manke96}.

Of all these approaches, a dual-state word-beam search for CTC decoding currently enjoys increased interest, \cite{Scheidl2018}, and will be described next. 
\begin{figure}[b!]
	\centering
	
		\includegraphics[scale=0.6]{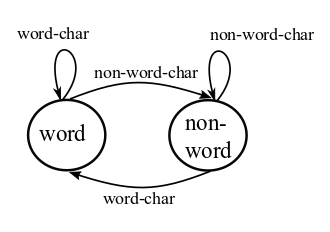}

	\caption{The dual-state word-beam search for CTC decoding\cite{Scheidl2018} used for our proposed system.}
	\label{wbs}
\end{figure}
\subsubsection{A dual-state word-beam search for CTC decoding} \label{wbs-sec}
The dual-state word-beam search for CTC decoding, \cite{Scheidl2018}, is based on Vanilla Beam Search Decoding (VBS) \cite{Hwang2016VBS} for decoding of the CTC layer. 
The output of RNN is a matrix, and it is the input of the dual-state word-beam search  method.
In the dual-state word-beam search, a prefix tree is made of groundtruth label of the training set. It consists of two states: word-state and non-word-state, Figure \ref{wbs}. The next character of the current beam is either a {\em word-character} or a {\em non-word-character}, and it determines the subsequent state of the beam.
The sets of {\em word-characters} and {\em non-word-characters} are predefined.

The temporal evolution of a beam depends on its state. For a beam in the non-word state, it is possible to be extended by a {\em non-word-character}, and it will stay in the non-word state. A word-character entering brings the system to the word state. Such a {\em word-character} is the beginning of a word. For a beam in the word-state the feasible following characters are presented by a prefix tree. This procedure iteratively repeats until a complete word is reached.

Scoring can be done in four ways:
\begin{enumerate}
\item Words: A dictionary is used without employing a language model (LM).

\item N-grams as LM: As a beam goes to non-word state from word state, the LM scores beam-labeling. 
\item Ngram+forecast: As a word-character appends a beam, prefix tree presents all possible words. LM scores all of the relevant beam-extensions.
\item Ngram+forecast+sample: to restrain the following potential words, first some samples are randomly selected. Then, LM scores them. The total score value has to be refined to account for the random-sampling step.
\end{enumerate}
The pseudo code of the dual-state word-beam search is illustrated in Algorithm \ref{algorithm1}. The list of symbols is as follows. 
\begin{itemize}
\item $RNN_{o}$: The sequence of RNN  output activations over time

\item $B$: the set of beams at the present time step.

\item $Width$: Beam width

\item $P$$_{b}$ : The probability of finishing the paths of a beam with blank.

\item $P$$_{nb}$ : The probability of not finishing the paths of a beam with blank.

\item $P$$_{tot}$: $P$$_b$+$P$$_{nb}$

\item $P$$_{txt}$: The probability allocated by the $LM$.

\item $T$: The final iteration of the algorithm, $t=T$.

\item $\O$: Empty beam.

\item $-1$: The last character of the beam.

\item $x$ : A beam.
\item $c$:  A character.
\item $x(t)$ : A beam character at t.
\item $numWords(x)$: the number of words in the beam $x$.
\item $GetbestBeams$($ B,\ Width$ ): Best $Width$ beams based on the highest value of $P_{txt}*P_{tot}$.

\item $NumWord's(x)$: The number of words exists in the beam $x$.
\item $scoreBeam(LM,x,c)$: The probability of seeing character $c$ for extension of the beam $x$.
\end{itemize}

\begin{algorithm}[t!]
\hrulefill

 \KwData{RNN output matrix $RNN$$_{o}$, $Width$ and $LM$}
 \KwResult{most probable labeling }
 $B$ = \{$\O$\}\;
 $P$ $_b$($\O$,0) $=$ 1\;

 \For  {$t =$  1 ...$T$}{
 $bestBeams$ $=$ $GetbestBeams$($B$,$Width$)
 
 $B$=$\{\}$\; 
 \For{$x$ $\in$ $bestBeams$}{

  \If{$x$ $!=$ $\O$}{
   $P$ $_{nb}$($x$,$t$) $+=$ $P$ $_{nb}$($x$,$t$$-$1)$*$ $RNN_{o}$($x$($-$1),$t$)\;
   }{}
    $P$ $_b$($x$,$t$) $+=$ $P$ $_{tot}$($x,t-$1)* $RNN_{o}$($blank$,$t$)\;
    $B$ $=$ $B$ $\cup$ $x$\;
    $nextChars$ $=$ $GetNextChars$($x$)\;
    \For  {$c$ $\in$ $nextChars$}{
    $x'= x + c$\;

    $P_{txt}(x') = scoreBeam(LM,x,c)$\;

    \eIf{$x(t) == c$}{
    $P_{nb}(x',t) += RNN_o(c,t)* P_b(x,t-1)$\;
    }{
    $P_{nb}(x',t) += RNN_o(c,t)* P_{tot}(x,t-1)$\;
    }    
    $B = B \cup x'$\;
    
        }
 }
 }
 $B$ $=$ $completeBeams$($B$)\;
 
\Return{$bestBeams$($B$,1)}\;
 
\hrulefill

$\ $

 \caption{The dual-state word-beam search for CTC decoding\cite{Scheidl2018}}\label{algorithm1}
\end{algorithm}

In RNNs, such as LSTM, the exact alignment of the observed word image with the ground truth label is not clear. Hence, a probability distribution at each time step is used for prediction. Which makes it more important to use an adequate coding scheme.

However, even after the CTC stage, additional processing steps from the above mentioned repertoire are needed to boost classifications.

Unfortunately, although, using linguistic resources is clearly advantageous, there are cases where this is not, or only partly possible:
\begin{itemize} 
\item Not all problems enjoy the presence of abundance or digitally encoded contemporary text content.
\item In historical collections there may be virtually no resources, not even a lexicon 
\item Many collections, e.g., administrative once have a dedicated jargon, abbreviations and non-standard phrasing. Even diaries may contain idiosyncratic neologisms.
\item Many collections have outdated geographical and scientific terminology, such as the historical document collection  belonged to Natuurkundige Commissie’s scientific exploration of the Indonesian Archipelago between year 1820 and 1850 \cite{Slovenia2017}. This heterogeneous handwritten manuscript contains 17,000 pages of the field notes based of the scientists'  natural observation in German, Latin, Dutch, Malay, Greek, and French. Biological terms vary greatly over periods in history \cite{Schuh2003}.
 
\end{itemize} 
There is, however, an additional way to improve the classification performance.
Impressive results using an ensemble method  were presented in  \cite{Stuner2016HandwritingRU}, however the number of networks was so large (118). that the need for a less drastic approach is becoming urgent. We will therefore focus on the probabilities of small-scale ensemble.

\subsection{Ensemble system}

A simple but effective method for improving an individual classifier performance is the ensemble method \cite{Ho1992,Ranawana2006,DBLP:journals/ijdar/GunterB03,Gunter2005,KarimiEtAk2015,YangEtAl2015,Menasri2012,kamHo1994,vanErp2000,Powalka1995-2}. In \cite{Ranawana2006}, it is shown that having diverse classifiers is a key point for classifier fusion.
Using ensembles for handwriting recognition with hidden-Markov models as basic word classifiers, \cite{DBLP:journals/ijdar/GunterB03} compares different ensemble creation methods: Bagging, AdaBoost, Half \& half
bagging, random subspace, architecture as well as different voting combination methods for handwriting-recognition task. It is shown than each of four methods, increases the performance.

The impact of dictionary size, the train-set size and the number of recognizers in ensemble systems is studied for off-line cursive handwritten-word recognition in \cite{Gunter2005}. The ensemble methods are Bagging, AdaBoost and the random subspace, while the recognizers are HMMs with different configurations. It is verified that increasing the size of the training set and the number of recognizers elevate the performance of the system, while the larger dictionary pull down the performance.

Recently, in \cite{KarimiEtAk2015}, ensemble classifiers for Persian handwriting recognition was used. They used AdaBoost and Bagging to combine weak classifiers created from hand-crafted families of simple features.

In the deep learning domain, \cite{YangEtAl2015} obtained very high accuracy for Chinese handwritten character recognition using deep convolutional neural networks and a hybrid serial-parallel ensembling strategy which tries to find an ``expert'' network for each example that can classify the example with a high accuracy, or if such a network cannot be found, falls back to the majority vote over all networks.

In \cite{Menasri2012}, an ensemble system is used for handwriting recognition of RIMES \cite{Rimes} dataset. The ensemble uses eight recognizers; including: Four variants of a recurrent neural network (RNN), a grapheme based MLP-HMM, and two variants of a context-dependent sliding window based on GMM-HMM. For RNN, a multi-dimensional long-short term memory neural network (MDLSTM) \cite{Graves2009} is used. 

In an ensemble system, majority voting can be used if the output of of each individual recognizer is only the best hypothesis label. If recognizers of ensemble system output a ranked hypotheses list, Borda count is possible \cite{kamHo1994,vanErp2000} to determine the result. In this case, it is required that the ranked list shows a sufficient  diversity of intuitive candidates, i.e., with a low edit distance from the target.
Two ensemble  system of handwritten recognition methods are presented in \cite{Powalka1995-2} : using word-list merging; and linear combination. 

The good results represented in literature are often based on a fairly complex system with many hyperparameters. In a e-science service such as Monk  which currently has about 530 different manuscripts, it is clear that human attendance and detailed selection of hyper parameters for each of those documents by human and crafting is impossible.
 
\section{Method}\label{Methodsec}

In this section, we present a limited-size ensemble system for word recognition with a minimum of human intervention. The suggested system uses an adequate label-coding scheme and a dictionary as the only resource for the language model. The system is described as follows.

\subsection{The Extra-separator coding scheme}
In the common coding scheme, we call it '\textit{Plain}', only the characters which are present in the word image appear in the corresponding label. In the '\textit{Extra-separator}' coding scheme, one more character is appended at the end of each label. The appended character, named the extra separator (e.g., '$|$'), must not exist in the alphabet of the dataset. The aim of adding the extra-separator character is to give the recognizer an extra hint concerning the end-of-word shape condition.

\subsection{Neural Network}
The neural network is a convolutional BiLSTM neural network, and it is an end-to-end trainable framework inspired by \cite{Shi2017AnET}. The main configuration of the networks is detailed in Table \ref{system}. In this section, we explain the essential components of our approach.

\begin{table}[t]
\caption{Configuration of our a convolutional recurrent neural network from input image (bottom) to last output (top). 'K', 'W','S' and 'P' denote kernel size, window size, stride and padding.}
\label{system}
\centering
\begin{tabular}{|cc|c|}
\hline 
Layer& &Configuration\\

\hline
$                   $&&A dual-state word-beam\\
Transcription&& search CTC decoding\\ 
\hline
$                   $&L1&512 hidden units\\
Bidirectional-LSTM &L2&512 hidden units\\
$                   $&L3&512 hidden units\\
\hline
Max Pooling&&W and S:1$\times$2\\
non-linear ReLU&&-\\
Normalization&&-\\
Convolution&&K:$3\times3$, S:1, p:1\\
\hline
Max Pooling&&W and S:1$\times$2\\
non-linear ReLU&&-\\
Normalization&&-\\
Convolution&&K:$3\times3$, S:1, p:1\\
\hline
Max Pooling&&W and S:1$\times$2\\
non-linear ReLU&&-\\
Normalization&&-\\
Convolution&&K:$3\times3$, S:1, p:1\\
\hline
Max Pooling&&W and S:2$\times$2\\
non-linear ReLU&&-\\
Normalization&&-\\
Convolution&&K:$3\times3$, S:1, p:1\\
\hline
Max Pooling&&W and S:2$\times$2\\
non-linear ReLU&&-\\
Normalization&&-\\
Convolution&&K:$3\times3$, S:1, p:1\\

\hline
\textbf{input image}  &&128$\times$32 gray-scale image\\
\hline
\multicolumn{1}{c}{\contour{black}{$\uparrow$}} \\
\end{tabular}
\end{table}
\subsubsection{Pre-processing}

The prepossessing is performed in each epoch of training. It is consists of:  a) data augmentation through randomly stretching/squeezing the gray-scale images in the width direction, b) re-sizing the images into $128 \times 32$ and c) normalization. 
Data augmentation is performed to increase the size of training set, and it is achieved by changing the width of an image randomly by a factor between 0.5 and 1.5.
Next, both the original gray-scale images and those added through data augmentation are re-sized so that either the width is 128 pixels or the height is 32 pixels. After that, we pad the image with white pixels until the size is $128 \times 32$. Then we normalize the intensity of the gray-scale image. 
Note that our method does not need baseline alignment or precise deslanting. Please note that one of our datasets was already deslanted to 90.

\subsubsection{A 5-layer CNN}

The pixel-intensity values after preprocessing are fed to the first of 5-layers of a CNN to extract feature sequences. Each layer of the CNN contains a convolution operation, 
normalization, the ReLU activation function \cite{Nair:2010:ReLU}, and a max pooling operation. 
The size of the kernel filters in each layer is $3 \times 3$. Given the fixed important hyperparameter setting, such as the number of layers, the only variable control parameters concern the number of units in the hidden layers. The simple table of three possible sizes $\{128,\ 256,\ 512\}$ is used with the random probability of 0.33 for selecting the sizes of hidden units. The sizes of the numbers of hidden units used in our experiments are shown in Table \ref{achsOfFeatureMaps}. The number of layers, size of kernel and optimizer is our configuration, and differ from \cite{Shi2017AnET}. 

 Furthermore, Instead of using ADADELTA \cite{Zeiler2012} used in \cite{Shi2017AnET}, we used RMSProp \cite{Tieleman2012}. Moreover, we used five convolutional layers instead of seven suggested in \cite{Shi2017AnET}. 
\begin{table}[t]
\caption{Number of hidden units in the CNNs front ends, in the five architectures used.}
\label{achsOfFeatureMaps}
\centering
\begin{adjustbox}{width=0.48\textwidth}
\begin{tabular}{|c|ccccc|}

\hline 

\multirow{2}{*}{\backslashbox{Arch.}{Layer}}&\multicolumn{5}{c|}{Hidden unit size}\\
\cline{2-6}
&\makebox[3em]{l$_1$}&\makebox[3em]{l$_2$}&\makebox[3em]{l$_3$} &\makebox[3em]{l$_4$}&\makebox[3em]{l$_5$}\\
\hline
$A_1$&128& 256& 256& 256& 512\\ 
$A_2$&128& 256& 512& 512& 512\\ 
$A_3$&128& 128& 256& 256& 512\\ 
$A_4$&128& 128& 512& 512& 512\\ 
$A_5$&128& 128& 128& 256& 512\\ 
\hline
\end{tabular}
\end{adjustbox}
\end{table}

\begin{table*}[b]
\caption{Datasets.}
\label{Datasets}
\centering
\begin{tabular}{|l|lll|lll|}

\hline

& \multicolumn{3}{c|}{RIMES}&  \multicolumn{3}{c|}{KdK} \\
 \cline{2-7}
set&Image&Word&Word CI&Image&Word&Word CI\\
\hline

Train(T)&51,738&4,943&4,639&103,464&8,717.6&8,006.8\\
Validation& 7,464&1,612&1,509&34,488&4,486.6&4,155.4\\
Test&7,776&1,692&1,606&34,488&4,486.6&4,155.4\\
\cline{1-7}
Whole dataset&66,978&5,744&5,378&172,440&11,749&10,747\\
\hline
\end{tabular}
\end{table*}

\subsubsection{BiLSTM}

The five convolutional layers are followed by three layers of BiLSTM. Because the last convolutional layer contains 512 hidden units, each BiLSTM has 512 hidden unit.

\subsubsection{Connectionist temporal classification (CTC)}
The CTC output layer contains two units more than characters in  the alphabet (A) of the given dataset: the suggested Extra separator (e.g., '$|$'), and  a common blank for CTC, which differs the space character. Therefore, the alphabet of CTC output is:

$A'=A\cup{extra\ separator\cup{blank}} $
The $|A+1|$ output units determine the probability of detecting the relevant label at the time. Further, the blank unit determines the probability of observing blank, or 'no label'. For CTC decoding, we use the dual-state beam search presented in \cite{Scheidl2018}. This method is explained in section \ref{wbs-sec}.

\subsection{The ensemble system}
For an input image, the outcome of the CTC decoder is  a string as a word hypotheses with its relative likelihood. The word hypothesis obtained from five networks are sent to the voter component. Plurality voting is then applied \cite{peleg1978}, where the alternatives are divided to subsets with identical strings. The subset 
with largest number of voters  are selected. In case of a tie, the subset with the highest averaged likelihood is the winner.
If the number of subsets is equal to the number of alternatives, the alternative with the highest likelihood is the winner. The winning string is considered as the final, best label of the input image. This approach was chosen after a pilot experiment, using Borda-count voting, whiteout good results. This may be due to the lack of diversity in the ranked candidate lists. Therefore,  the more simple approach using plurality voting with exception handling was  performed.

\section{Results} \label{resultsec}
In this section, firstly, we describe the datasets used in the experiments. Then, we explain how our experiments were carried out. Finally, we report the numerical results.

\subsection{Datasets}

In this paper, we used two datasets which differ in time period and language, summarized in Table \ref{Datasets}.  The first dataset is named  RIMES, which was used to be comparable with the state-of-the-art methods. This database has different versions. We used isolated words of the version of ICDAR 2011 for evaluation of the methods and making the comparison with the published results possible \cite{grosicki:RIMES}. 
The RIMES database is drawn from different types of handwritten manuscripts: postal mails and faxes. It contains 12,723 pages written by 1,300 volunteers using  black ink on white paper. 
The RIMES dataset consists of 51,738 images of French handwriting for training, 7,464 images for validation and 7,776 images for testing. The dictionary size of the training set is 4,943 words, the validation set is 1,612 and the test set is 1,692, and the dictionary size of the whole dataset is 5,744 words. The comparison is accomplished case insensitive as it is common for the RIMES dataset, and the accent were contemplated. In the evaluation process of our model on RIMES, two dictionaries were used:  {\em Concise} and {\em Large}. The {\em Concise} dictionary contains the whole words within the RIMES dataset, $n_{words}$ = 5,744\ (6K). A French dictionary called {\em Large} (50K) is used to study the effect of a larger dictionary.

 \begin{figure}[b]
   	\centering
	\subfigure[garnizoensplaats$|$]{\includegraphics[scale=0.14]{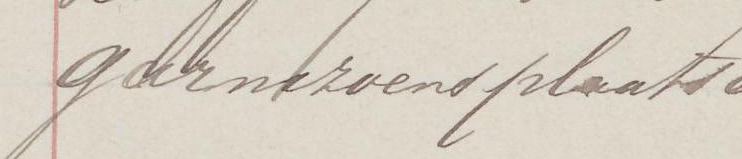}}\\
	\subfigure[advocaat$|$]{\includegraphics[scale=0.14]{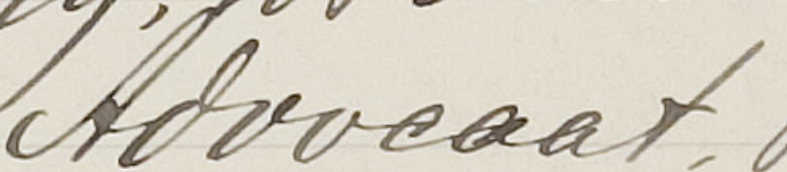}}\\	
		\subfigure[Staatsblad$|$]{\includegraphics[scale=0.14]{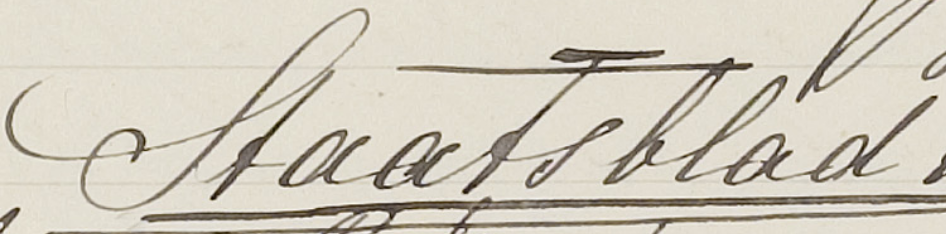}}\\
	\subfigure[wetenschappelijken$|$]{\includegraphics[scale=0.14]{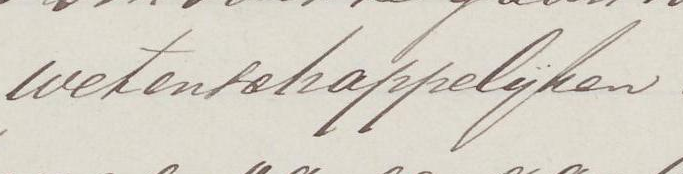}}
	\caption{Samples of the KdK dataset (the year 1903). (a) to (d) show the images labeled, using the Extra-separator coding scheme.}
	\label{KdK}
\end{figure}
The second dataset belongs to the National Archive of the Netherlands, named KdK (Het Kabinet der Koningin or Dutch Queen’s Office)\cite{Zant2008,VANOOSTEN20141031}. The manuscript was written between years 1798 and 1988, the year 1903 was used. The KdK dataset contains 172,440 word images. The number of word classes of the total dataset is 11,749 and 10,747, case-sensitively and case-insensitively, respectively. Regardless of case-sensitivity, there are 1 to 5,628 sample(s) in each class. The length of  the word samples is 1 to 28 character.  In the case-sensitive manner, 5\% of the test words does not occur in the training words, and is 'out of vocabulary (OOV)'. OOV in the case-insensitive manner, is 4.5\%. The remaining words are considered as is referred to as 'in vocabulary (INV)'.  Figure \ref{KdK} shows four original samples of the KdK dataset. For evaluation, two dictionaries are used: {\em Concise} and {\em Large}. The {\em Concise} dictionary contains all the words in the KdK dataset (12K); the size of the Dutch {\em Large} dictionary is 384K, \cite{WOORDEN}.

\subsection{Quantitative results}

\begin{table*}[t]

\caption{The result of the RIMES dataset. The Table shows comparison of word accuracy (\%) between two coding schemes (Plain and Extra separator) using Best-path CTC decoder and the dual-state word-beam search with different dictionary sizes in terms of average $\pm$standard deviation (avg $\pm$sd) and The ensemble. Two dictionaries is used; the dictionary  contains words of the train, the validation and the test sets (Concise) and a dictionary which contains more than 50K words (Large).}
\label{ComparisonFor-Bar-majority}
\centering
\begin{adjustbox}{width=\textwidth}
\begin{tabular}{|l|llll|lllr|}

\hline
Coding scheme& \multicolumn{4}{c|}{Plain}& \multicolumn{4}{c|}{ Extra separator} \\ 

\hline

CTC decoder&Best path&& \multicolumn{2}{c|}{ The dual-state word-beam search}&Best path&& \multicolumn{2}{c|}{ The dual-state word-beam search}\\
\cline{1-1}\cline{2-2}\cline{4-6} \cline{8-9}
\backslashbox{Arch.}{Dictionary}&-(dictionary-free)&&Concise (6K)&Large (50K)&-(dictionary-free)&&Concise (6K) &Large (50K)\\

\hline
A$_1$&84.6&&94.1&92.9&83.8&&95.2&94.1$\ \ \ \ \ \ \ $\\ 
A$_2$&84.6&&94.5&93.2&84.4&&94.7&93.4$\ \ \ \ \ \ \ $\\ 
A$_3$&84.2&&94.4&93.1&84.9&&95.5&94.6$\ \ \ \ \ \ \ $\\ 
A$_4$&84.5&&94.2&92.8&84.9&&95.3&94.3$\ \ \ \ \ \ \ $\\ 
A$_5$&84.7&&94.2&92.6&84.3&&95.1&94.0$\ \ \ \ \ \ \ $\\
\hline
avg $\pm$sd&84.5 $\pm$0.2&&94.3 $\pm$0.2&92.9 $\pm$0.2&84.5 $\pm$0.4&&95.2 $\pm$0.3&94.1 $\pm$0.4\\ 
\hline
Ensemble&88.4&&95.7&94.8&88.9&&96.6&95.8$\ \ \ \ \ \ \ $\\
\hline
\end{tabular}
\end{adjustbox}
\end{table*}

In this section, we evaluate our model on the RIMES and the KdK datasets in terms of coding scheme (Plain vs Extra separator) and ensemble/single network. Moreover, for the RIMES dataset, the results of our model is compared with the-state-of-the-art methods suggested in \cite{Stuner2016HandwritingRU,Stuner2017,Poznanski,Menasri2012,Ptucha2019,Sueiras2018}. In \cite{Dutta2018}, very good result are reported. However, their system was trained with a large amount of synthetic data. Therefore, we do not find it comparable with our approach, which is exclusively based on the the given dataset, and its augmentations. 

For the Extra-separator coding scheme, a character which is absent in the given dataset was found automatically as the extra-separator character, the bar sign ($|$); hence,the bar sign is annexed to the end of each image label, Figure \ref{KdK}. As a result, the size of the output of the CTC layer increases. The RIMES dataset contains 80 unique characters. Meaning that the size of the output layer of the CTC layer is 82 (80 unique character, one extra separator, and one common blank). The KdK dataset contains 52 unique characters. Therefore, the size of output layer of CTC layer is 54 (52 unique character, one extra separator, and one common blank). We compare the result of this addition to the Plain coding scheme. Two CTC decoder methods are used: dictionary-free (Best path) and with dictionary (dual-state word-beam search \cite{Scheidl2018}). For the dual-state word-beam search, two dictionaries are used for each dataset; Concise and Large.

Table \ref{ComparisonFor-Bar-majority} shows the effect of the two coding schemes, single recognizer and ensemble voting on the RIMES dataset showing word accuracy (\%). For each of the two coding schemes (Plain and Extra separator), the five architectures were trained, which resulted to 10 trained networks. Then the networks were evaluated using  the Best-path CTC decoder and the dual-state word-beam search CTC decoder applying the Concise (6K) and the Large (50K) dictionaries. The result of each evaluation and the relative average $\pm$ standard deviation (avg $\pm$sd) are reported. In the bottom row of the Table \ref{ComparisonFor-Bar-majority}, the voting-based result of the ensemble of the five networks is presented. 
 
{\em Best path vs Dual-state word-beam search}: the results confirm that using a decoder with dictionary considerably improves the performance (95-97\%) as expected (t-test, $p < 0.05$, significant). The dictionary-free Best-path CTC decoder is given a low performance, still at 88-89\%. Moreover, when the dual-state word-beam search CTC decoder is used, adding an extra-separator character  enhances the model.  

{\em Plain vs Extra separator}: for the Best-path CTC decoder, both Plain and Extra separator have an average of 84.5\%, (t-test, $p > 0.05$, N.S.). Therefore, the extra separator has no effect. However, for a dual-state word-beam search CTC decoder using the Concise dictionary, Plain has an average of 94.3\%, and Extra separator has an average of 95.2\%, (t-test, $p < 0.05$, significant). Hence, the extra separator is effective; for a dual-state word-beam search CTC decoder, using the Large dictionary, Plain has an average of 92.9\%, and Extra separator has an average of 94.1\%, (t-test, $p < 0.05$, significant).
Therefore, the Extra separator is effective again, for the case of a large dictionary. 

{\em Single network vs Ensemble}: ensemble voting increases the performance where its effect is more on a weaker recognizer (4 pp increase in performance for the dictionary-free CTC decoder using the Plain/Extra separator coding scheme, final row vs average and individual). An ensemble of five recognizers, using the CTC decoder with the Concise dictionary combined with the Extra-separator coding scheme results in the highest performance (96.6\%, column 6, bottom).

\begin{table}
\caption{The comparison of our system to the state-of-the-art systems on the RIMES dataset in terms of number of recognizers (n$_{rec}$), homogeneity of the algorithm (Hom.), complexity of the approach (Compl.), and word accuracy (\%) (word$_{acc}$).  Please, refer to the text for further explanation.}
\label{others}
\centering
\begin{adjustbox}{width=0.48\textwidth}
\begin{tabular}{|c|cccc|}
\hline 
system &n$_{rec}$&Hom.&Compl.&word$_{acc}$\\
\hline
\multicolumn{1}{|l|}{Ours} &1&N/A&low&95.1 $\pm$0.3\\
\multicolumn{1}{|l|}{Ours (Table \ref{achsOfFeatureMaps})}&5&\checkmark&low&96.6\\
\hline
\multicolumn{1}{|l|}{Stuner et al. 2016 \cite{Stuner2016HandwritingRU}}&2,100&\xmark&medium&96.5\\
\multicolumn{1}{|l|}{Stuner et al. 2016 \cite{Stuner2016HandwritingRU}}&118&\xmark&medium&96.4\\
\multicolumn{1}{|l|}{Poznanski and Wolf 2016 \cite{Poznanski}}&1&N/A&medium&96.1\\
\multicolumn{1}{|l|}{Menasri et al. 2012 \cite{Menasri2012}}&8&\xmark&high&95.3\\
\multicolumn{1}{|l|}{Ptucha et al. 2019 \cite{Ptucha2019}} &3&\xmark&high&94.3\\

\multicolumn{1}{|l|}{Menasri et al. 2012 \cite{Menasri2012}}&1&N/A&low&91.1\\

\multicolumn{1}{|l|}{Stuner et al. 2017 \cite{Stuner2017}}&1&N/A&low&89.9\\
\multicolumn{1}{|l|}{Sueiras et al. 2018 \cite{Sueiras2018}}&1&N/A&low&86.9\\
\hline

\end{tabular}
\end{adjustbox}
\end{table}

 \begin{figure}[b!]
	\centering
	
		\includegraphics[width=0.48\textwidth]{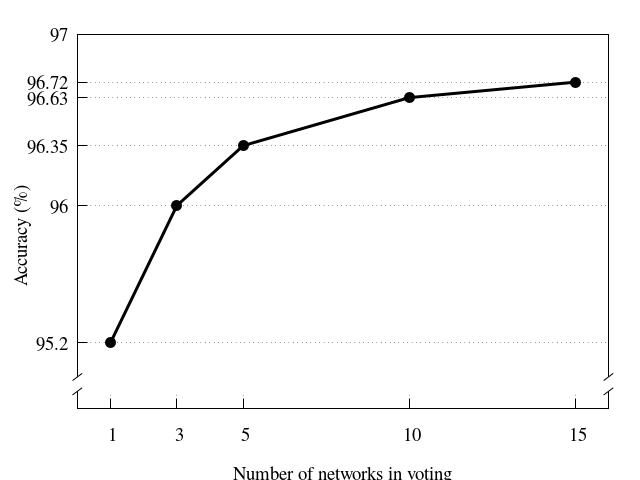}
		\caption{The graph shows the effect of number of networks in the voting ensemble on the final accuracy (\%) for the RIMES dataset, with diminishing returns as the number of voters increases.}
		\label{growing}
		\end{figure}

To study the effect of the number of networks in the ensemble on the final accuracy, the result of randomly selected 1, 3, 5, 10 and 15 network(s) are shown in Figure \ref{growing} for the RIMES dataset. The coding scheme is Extra separator, and CTC decoder is the dual-state word-beam search using the Concise dictionary. The networks in the ensemble only differ in the random initialization and number of the units over the layers, also randomly selected from the set $ n = \{128,\ 256,\ 512\}$ in 1 through 4. The maximum accuracy is obtained by the ensemble of 15 networks, 96.72\%, which is just 0.09 pp is more than using 10 networks.

Table \ref{others} shows the comparison of our method on the RIMES dataset with \cite{Stuner2017,Stuner2016HandwritingRU,Poznanski,Menasri2012,Sueiras2018,Ptucha2019}  in the terms of a number of characteristics: number of recognizers, homogeneity of the algorithm, word accuracy (\%) and the complexity of the approach, not to be confused with computational complexity, e.g. deep learning method without extra complicated modules.

\begin{table} [b]

\caption{The results of the KdK dataset. The Table shows the average (avg) and standard deviation (sd) of word accuracy (\%) of five architectures using 5-fold cross-validation and varying per architecture, over the following  parameters: dictionary (none, Concise, Large)and coding scheme (Plain, Extra separator) ($5\times3\times2$). Each row is derived from 30 network evaluations.}

\label{Comparison-KdK-Arc-dic-labelling}
\centering
\begin{tabular}{|l|lr|}
\hline
Arch&avg&sd\\

\hline
A$_1$&94.4&2.7\\ 
A$_2$&94.4&2.6\\ 
A$_3$&94.3&2.8\\ 
A$_4$&94.4&2.6\\ 
A$_5$&94.2&2.8\\ 
\hline
\end{tabular}

\end{table}

\begin{figure}[tp]
	\centering
	
		\includegraphics[width=0.45\textwidth]{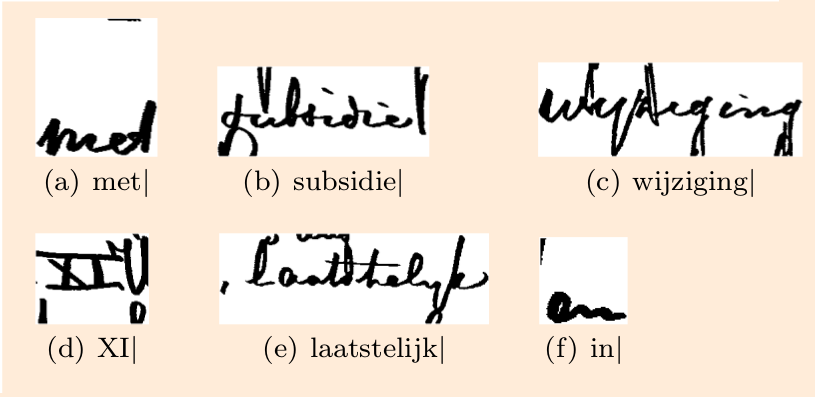}
		\caption{The samples of the pre-processed KdK dataset. (a) to (f) show the images labeled using the Extra-separator coding scheme. After the binarization process, all the word images were sheared 45 degrees in the anti-clockwise direction.  }
	
	\label{KdK-removed-white}
		\end{figure}

For the KdK dataset, the results are as follows. The samples of the KdK dataset for our experiment were binarized, then sheared 45 degrees in the anticlockwise direction to the slant angle in this style approximately 45 degrees. Afterwards, the white borders of images were removed horizontally and vertically, until the place where the first black pixel is observed. In Figure \ref{KdK-removed-white} the deslanted, white-removed images are shown. 
 To derive a more accurate estimation of the performance of our model, we ran 5-fold cross-validation. Each architecture, $A_i,$ where $i\ =\ 1\ to\ 5$, is trained, either using the Plain coding scheme or using the Extra separator, resulting in 50 trained networks ($5\times5\times2$). Then, each network is tested three times: using the dictionary-free Best-path CTC decoder; using the dual-state word-beam search CTC decoder applying Concise (12K) and Large (384K).

  Table \ref{Comparison-KdK-Arc-dic-labelling} shows the average (avg) and standard deviation (sd) of word accuracy (\%) of five architectures using 5-fold cross-validation and varying per architecture, over the following  parameters: dictionary (none, Concise, Large), and coding scheme (Plain, Extra separator) ($5\times3\times2$). Each row is derived from 30 network evaluations. In other words, each row is the result of one architecture, regardless of the used CTC decoding method, dictionary, and coding scheme. Slightly lower performance is expected as the Best-path CTC decoder pulls it down. Similar result is achieved for each coding scheme,  regardless of  the used CTC decoding method, dictionary, and architecture. The Extra-separator has a higher performance, 94.5\%, which is 0.4 pp higher than the Plain decoding scheme. 
 
\begin{figure*}[bp]
	\centering
	
		\includegraphics[width=0.8\textwidth]{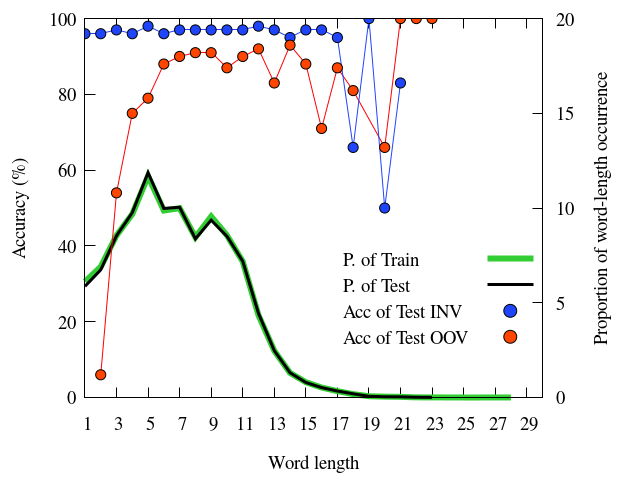}
	\caption{The continuous lines indicate the word length proportion of the train and the test set of one round of 5-fold cross-validation for the KdK dataset. The dots represent the accuracy of the network A$_2$ on OOV and INV using the dual-state beam search and an extra separator for labeling. Please note that OOV words can be recognized with a high accuracy in a range which there are few number of samples in the test set.}
	\label{length-acc}
\end{figure*}

The Table \ref{Comparison-KdK-Arc-dic-labelling-2} shows the average (avg) and standard deviation (sd) of word accuracy (\%) of using dictionary on 5-fold cross-validation and varying per dictionary, over the following  parameters: architecture (A$_i,\ i= 1$ to $5$), coding scheme (Plain, Extra separator) ($5\times5\times2$). Each row is derived from 50 network evaluations.

\begin{table}

\caption{The results of the KdK dataset. The Table shows the average (avg) and standard deviation (sd) of word accuracy (\%) of using dictionary on 5-fold cross-validation and varying per dictionary, over the following  parameters: architecture (A$_i,\ i= 1$ to $5$), coding scheme (Plain, Extra separator) ($5\times5\times2$). Each row is derived from 50 network evaluations.}

\label{Comparison-KdK-Arc-dic-labelling-2}
\centering
\begin{adjustbox}{width=0.48\textwidth}
\begin{tabular}{|l|l|lr|}

\hline
CTC decoder&Dictionary&avg&sd\\

\hline
\multirow{2}{*}{Dual-state word-beam search}&Concise (12K)&96.5&0.3\\
&Large (384K)&95.8&0.4\\
\hline
Best-path method&dictionary-free&90.6&0.3\\
\hline
\end{tabular}
\end{adjustbox}
\end{table}

Figure \ref{length-acc} shows the behavior of a single network A$_2$, using the Extra-separator coding scheme and the dual-state word-beam search CTC decoder. For different word lengths and for the OOV and INV condition in the KdK dataset. The blue and red dots represent the accuracy on OOV and INV, respectively.  
\begin{figure}[t]
	\centering
	
		\includegraphics[width=0.5\textwidth]{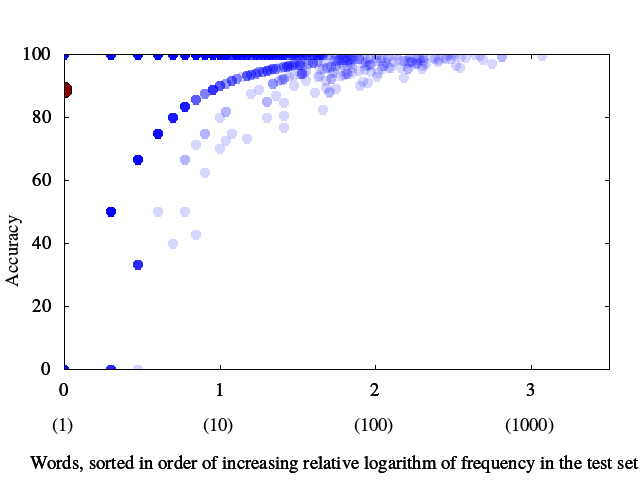}
	\caption{Accuracy of words obtained by network $A_2$ on one round of 5-fold cross-validation on the KdK dataset. The $X$ axis represents words, sorted in order of increasing frequency ($f$) in the test set ($\#$samples$=34,488$). The parentheses shows the number of sample per class. The blue circles show the words which are present in the training set, in vocabulary (INV), where the darker the blue circle, the more word classes. The dark red circle indicates the average accuracy of out-of-vocabulary samples (OOVs) at $f=0$.	 }
	\label{plot-acc}
\end{figure}

\begin{table*}[b]

\caption{The result of the KdK dataset. The Table shows comparison of word accuracy (\%) between two coding schemes (Plain and Extra separator) using Best-path CTC decoder and the dual-state word-beam search with different dictionary sizes in terms of average $\pm$standard deviation (avg $\pm$sd) and The ensemble. Two dictionaries is used; the dictionary  contains words of the train, the validation and the test sets (Concise) and a dictionary which contains 384K words (Large).}
\label{KdK-majority-table}
\centering
\begin{adjustbox}{width=\textwidth}
\begin{tabular}{|l|llll|lllr|}

\hline
Coding scheme& \multicolumn{4}{c|}{Plain}& \multicolumn{4}{c|}{ Extra separator} \\ 

\hline

CTC decoder&Best path&& \multicolumn{2}{c|}{ The dual-state word-beam search}&Best path&& \multicolumn{2}{c|}{ The dual-state word-beam search}\\
\cline{1-1}\cline{2-2}\cline{4-6} \cline{8-9}
\backslashbox{Arch.}{Dictionary}&-(dictionary-free)&&Concise (12K)&Large (384K)&-(dictionary-free)&&Concise (12K) &Large (384K)\\

\hline
A$_1$&90.62&&96.27&95.52&90.84&&96.79&96.16\\
A$_2$&90.77&&96.34&95.55&90.87&&96.76&96.15\\
A$_3$&90.45&&96.23&95.43&90.50&&96.77&96.13\\
A$_4$&90.72&&96.32&95.56&90.88&&96.81&96.20\\
A$_5$&90.23&&96.13&95.34&90.50&&96.72&96.09\\
\hline				
avg $\pm$sd&90.56 $\pm$0.22&&96.26 $\pm$0.12&95.48 $\pm$0.15&90.72 $\pm$0.26&&96.77 $\pm$0.12&96.14 $\pm$0.13\\

\hline
Ensemble&93.38 $\pm$0.12&&97.00 $\pm$0.09&96.51 $\pm$0.10&93.52 $\pm$0.13&&97.37 $\pm$0.09&96.99 $\pm$0.11\\						

\hline
\end{tabular}
\end{adjustbox}
\end{table*}
The continuous green and black lines in Figure \ref{length-acc} indicate the word-length occurrence of the train and the test sets in the KdK dataset in one round of the 5-fold cross-validation. It is demonstrated that the single network $A_2$ on INV words with a length up to 17 characters has a high accuracy and is promising. For longer words the performance becomes erratic.
The single network $A_2$ does not perform satisfactorily on short OOV words with 1 to 4 characters. The performance on OOV words which have 5 to 15 characters is highly adequate. For OOV words whose length is between 16 and 20, the performance is variable. Surprisingly, for OOV samples longer than 21 characters, the model has a high performance.

Figure \ref{plot-acc} shows the accuracy of words achieved by network A$_2$ on one round of 5-fold cross-validation on the KdK dataset. On the $X$ axis words are sorted in order of increasing relative log frequency of the test set. The blue circles indicate INV words. The dark red circle reveals the average accuracy and the log occurrence of OOV. Note the different '\textit{threads}' in the curve, revealing groups of easy and difficult (slow-starting) classes. In a lifelong machine-learning, the horizontal axis corresponds to time, starting with just a few examples on the left. The average of the performance on OOV samples is high, at $log(f) = 0$, where $f$ is frequency in the test set ($\#$samples$=34,488$).

Table \ref{KdK-majority-table} shows the comparison of the effect of the two coding schemes (Plain and Extra separator) and the CTC decoder application on the ensemble for the five rounds of the cross-validation of the KdK dataset.
  
{\em Best path vs Dual-state word-beam search}: using no dictionary conditions in more than 93\% accuracy. Using a decoder with dictionary boosts the performance (t-test, $p < 0.05$, significant). Adding an extra separator enhances the model, when a CTC decoder with dictionary is used.

{\em Plain vs Extra separator}: for the Best-path CTC decoder, Plain has an average of 90.6\%, and Extra separator has an average of 90.7\% (t-test, $p > 0.05$, N.S.). Therefore, the extra separator has no effect ; for a dual-state word-beam search CTC decoder using the Concise dictionary (12K), Plain has an average of 96.3\%, and Extra separator has an average of 96.8\% (t-test, $p < 0.05$, significant). Therefore, the extra separator is effective ; for a dual-state word-beam search CTC decoder using the Large dictionary (384K), Plain has an average of 95.5\%, and Extra separator has an average of 96.1\% (t-test, $p < 0.05$, significant). Therefore, the extra separator is effective .

{\em Single network vs Ensemble}: ensemble voting increases the performance where its effect is more on a weaker recognizer (3 pp increase in performance for the dictionary-free CTC decoder for Plain/Extra separator). Ensemble of five recognizers used the CTC decoder with the Concise dictionary combined with the Extra separator coding scheme results in the highest performance (97.4\%).

\begin{figure}[t] 

	\centering
	\includegraphics[width=0.5\textwidth]{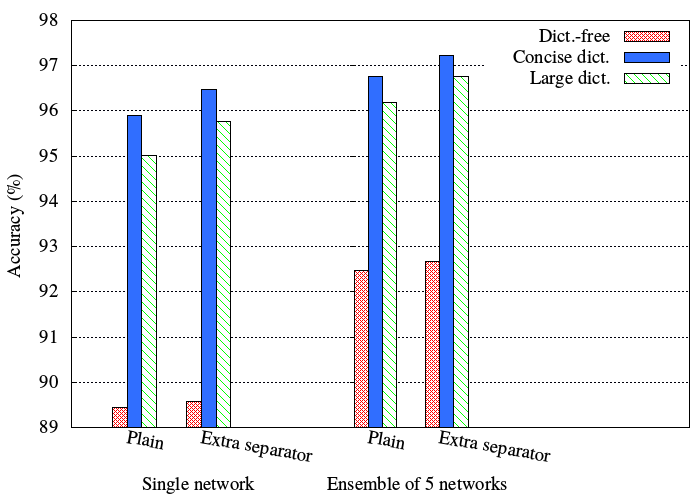}
	\caption{ Comparison of the effect of the two coding schemes (Plain vs Extra-separator) and dictionary application on the single architecture and ensemble voting on the RIMES and the KdK datasets showing the weighted average based on test set sizes. The two datasets are rather different. The spread of a distribution is not very informative.}
	\label{KdK-RIMES}
\end{figure}

Figure \ref{KdK-RIMES} shows comparison of the effect of two coding schemes and dictionary application on single architecture and ensemble voting on the RIMES and the KdK datasets showing the weighted average.
Table \ref{KdK+RIMES-t} shows the average of word accuracy (\%) on the RIMES and KdK datasets, using the Concise dictionary and the Extra-separator coding scheme.

 \begin{table} [t]
\caption{ Weighted average of word accuracy (\%) on the RIMES and KdK datasets, using the dual-state word-beam search applying the Concise dictionary and the Extra-separator coding scheme, for the two CTC methods and single vs ensemble voting}
\label{KdK+RIMES-t}
\centering
\begin{adjustbox}{width=0.48\textwidth}
\begin{tabular}{|l|lr|}
\hline
\backslashbox{CTC decoder}{Framework}&Single&\multicolumn{1}{l|}{Ensemble}\\
\hline
Best path&89.6&92.7\\ 
Dual-state word-beam search&96.5&97.3\\ 
\hline
\end{tabular}
\end{adjustbox}
\end{table}

\section{Discussion} \label{discussionsec}
The results indicate that it is possible to achieve a high word accuracy (\%) in comparison to the state of the art with a limited-size ensemble, a homogeneous algorithmic approach and a low complexity  \cite{Stuner2016HandwritingRU,Stuner2017,Poznanski,Menasri2012,Sueiras2018,Ptucha2019} (cf. Table \ref{others}). In those studies, numerous networks (up to 118 or 2100 network instances) are required in the ensemble. Whereas our method only uses five networks, yielding comparable or better results. In the proposed method, feature descriptors such as histogram of oriented gradients (HOG~\cite{Dalal}) are not used, the process starts with a pixel image and is trained end to end.

Results also indicate that the average performances of the two coding schemes (Plain and Extra separator) differ significantly if the dual-state word-beam search is used for CTC decoding. 
 In other words, the extra-separator character, '$|$', tagging the end of the word, boosts the result of the dual-state word-beam search CTC decoding. This increase in performance occurs despite the slight increase of the model size by adding the extra-separator character. However, the effect on the result of CTC best-path decoding, i.e., a non-dictionary method, is limited. On the other hand, using the decoder with dictionary boosts the performance. Finally, ensemble voting clearly improves the word accuracy (\%); its effect is stronger on weaker recognizers.

It should be noted that the reported result is based on realistic images with many word-segmentation problems, and therefore can be considered as a conservative estimate ( cf. Figure \ref{KdK-removed-white}).

We have shown that medium length OOV words (5 to 11 characters) profit from training information in the shorter words in the training set (cf. Figure \ref{length-acc}). Longer OOV words (11 to 23 characters) profit from the training on words whose length is 1 to 11 characters. Interestingly, OOV can have a high performance in a range for which there are not many examples (cf. Figure \ref{length-acc}). In addition, for INV words shorter than 18 characters, the accuracy is higher than 95\%. Therefore, our method recognized the common length OOV and INV words with a high accuracy. Alternatively stated, we demonstrate an important finding on a single network where increasing the size of difficult in-vocabulary word classes  yields superior results, while the performance on easy in-vocabulary word classes is high even for a limited number of samples.

The goal of this research is not a record attempt towards maximized accuracy on the RIMES and the KdK datasets. Higher performance can undoubtedly be achieved using a larger ensemble. However, our choice for an ensemble of 5 voting elements results in a compromise with a very good and stable performance. The more than 1 pp jump in performance from one individual classifier to five classifiers is larger than the less than 0.3 pp increase in performance from 5 to 10 classifiers, and the increase in the performance is even smaller for higher numbers of classifiers in the amble, showing diminishing returns.

Furthermore, we have shown that by providing a more than 30 times larger dictionary, only a slight drop in performance occurred.
In addition, for the dictionary-free approach, using an ensemble system results in a much higher performance with more stability than a single network. 
 However, in the higher performing approach, the relative improvement is present but less prominent, when a dictionary is used. Moreover, as expected from previous research, using the CTC decoder with a dictionary increases the performance of our model compared to dictionary-free CTC decoder.

\section{Conclusions} \label{conclusionsec}
This study was aimed at achieving high-performance handwritten word recognition, using deep learning, however, with a limited cost in terms of network handcrafting combined with low complexity. Our model consists of an ensemble of just five homogeneous end-to-end trainable recognizers, using plurality voting with a solution for ties. Each recognizer is composed of five convolutional layers and three BiLSTM layers, followed by a CTC layer. Diversity is fostered by various number of units in the hidden layers of the CNNs. For CTC decoding, a dual-state word-beam search is applied, using only the given dictionary as the only language model. Furthermore, we study the effects of the dictionary-free Best-path CTC decoding on a single network and on the ensemble. Training the system is done from scratch, exclusively on the given dataset, and data augmentation is not used during testing. The word accuracy (\%) of our model is 96.6\% on RIMES, and 97.4\% on the KdK dataset, a locally collected historical handwritten dataset. Results show that an ensemble size higher than five networks only yields limited further improvement; the method is not very sensitive to diverse network correspondence. Moreover, we showed that using an extra separator in the label-coding scheme boosts the performance with advantage of using it in case of a large dictionary.  
 
We showed that by providing $\sim 30$ times larger dictionary, only a slight drop in performance occurred. Ensemble voting improves the performance; its effect is more on weaker recognizers.
Longer out-of-vocabulary (OOV) words benefit from training information in the shorter words in the training set.

On in-vocabulary word classes, increasing the number of samples yields better results. However, it does not have an effect on easy word classes. The performance of our model is even relatively high for OOV classes in word-length ranges, where there are a limited number of samples in the training set. The suggested method is applicable to e-Science services where it is not feasible to manually tailor hyperparameters, pre-processing and language model for each manuscript based on prior knowledge. 

Word-based LSTMs cannot make use of the large textual content. Therefore, as future work, we plan to extend our approach to handle the handwritten line recognition task. Moreover, we will explore the applicability of our model on other datasets with different languages, and increase the performance on out-of-vocabulary words. Furthermore, the challenge of high-performance recognition of long words will be addressed.

\section*{Acknowledgment}

This work is part of the research programme Making Sense of Illustrated Handwritten Archives with project number 652-001-001, which is financed by the Netherlands Organisation for Scientific Research (NWO). We would like to thank Gideon Maillette de Buy Wenniger for thoughtful advice and the Center for Information Technology of the University of Groningen for providing access to the Peregrine high performance computing cluster.

\bibliographystyle{spbasic}      
\bibliography{ms.bbl}

\end{document}